\documentclass[default,iicol]{sn-jnl}

\jyear{2021}%

\usepackage{amsmath}

\theoremstyle{thmstyleone}%
\theoremstyle{thmstyletwo}%

\theoremstyle{thmstylethree}%

\begin{document}

\title[Article Title]{Design and Operation of Autonomous Wheelchair Towing Robot}


\author[1]{\fnm{Hyunwoo} \sur{Kang}}\email{gusdn91t@naver.com}
\equalcont{These authors contributed equally to this work.}

\author[1]{\fnm{Jaeho} \sur{Shin}}\email{tlswogh423@naver.com}
\equalcont{These authors contributed equally to this work.}

\author[1]{\fnm{Jaewook} \sur{Shin}}\email{sjw4689@naver.com}

\author[2]{\fnm{Youngseok} \sur{Jang}}\email{duscjs59@gmail.com}

\author*[1]{\fnm{Seung Jae} \sur{Lee}}\email{seungjae\_lee@seoultech.ac.kr}

\affil*[1]{\orgdiv{Department of Mechanical System Design Engineering}, \orgname{Seoul National University of Science and Technology}, \orgaddress{\street{Gongrung-ro 232}, \city{Seoul}, \postcode{01811}, \country{Republic of Korea}}}

\affil[2]{\orgdiv{Department of Aerospace Engineering}, \orgname{Seoul National University}, \orgaddress{\street{Gwanak-ro 1}, \city{Seoul}, \postcode{05116}, \country{Republic of Korea}}}


\abstract{
In this study, a new concept of a wheelchair-towing robot for the facile electrification of manual wheelchairs is introduced. 
The development of this concept includes the design of towing robot hardware and an autonomous driving algorithm to ensure the safe transportation of patients to their intended destinations inside the hospital.
We developed a novel docking mechanism to facilitate easy docking and separation between the towing robot and the manual wheelchair, which is connected to the front caster wheel of the manual wheelchair. 
The towing robot has a mecanum wheel drive, enabling the robot to move with a high degree of freedom in the standalone driving mode while adhering to kinematic constraints in the docking mode. 
Our novel towing robot features a camera sensor that can observe the ground ahead which allows the robot to autonomously follow color-coded wayfinding lanes installed in hospital corridors. 
This study introduces dedicated image processing techniques for capturing the lanes and control algorithms for effectively tracing a path to achieve autonomous path following. 
The autonomous towing performance of our proposed platform was validated by a real-world experiment in which a hospital environment with colored lanes was created.
A video demonstration of our robot can be found at \url{https://www.youtube.com/watch?v=RdnuhTjrMQc}.
}




\maketitle

\section{Introduction}

Many patients with mobility difficulties in hospitals rely on wheelchairs for locomotion.
However, traditional manual wheelchairs require patients to use their arm strength to propel, making it difficult for those with physical impairments to move independently.
One possible solution to overcome this problem is to introduce electric-powered wheelchairs (EPWs) \cite{electric-powered_wheelchairs}, which utilize electric motors and batteries to bring motion to the wheelchair.
However, this solution is cost-prohibitive since it involves replacing manual wheelchairs already equipped in hospitals with expensive EPWs.
Additionally, most existing EPWs have a non-foldable design and occupy much space, making storage challenging.

Furthermore, healthy patients who can operate manual wheelchairs may still face significant challenges when moving around hospitals.
Hospitals typically have complex corridor structures, making it difficult for patients unfamiliar with the environment to navigate to their destination.
To address this issue, many hospitals install color-coded wayfinding lanes on the floor to allow patients to follow a specific color-coded lane to reach their destination \cite{color_lane}.
However, for patients with co-morbidities such as cognitive impairment in addition to reduced physical strength, the wayfinding lanes may not be sufficient for them to arrive at their destination independently.

\subsection{Contribution}
In this paper, we introduce a new engineering solution to the challenges associated with manual wheelchair transportation within hospitals by developing a new wheelchair-towing robot platform. 

The proposed robot hardware is designed to operate in both standalone and towing modes. 
In standalone mode, the robot moves by itself to a specified location within the hospital upon request, without a wheelchair. 
Once it arrives, the robot enters towing mode and docks with manual wheelchairs to tow the wheelchair to its destination.
By enabling hospitals to electrify manual wheelchairs selectively, the proposed approach eliminates the need to equip electrification tools for each and every manual wheelchair, thereby reducing the overall cost of implementing such a solution.

In addition to the robot hardware design, this study also presents a smart wheelchair system \cite{smart_wheelchair, smart_wheelchair2, smart_wheelchair3, smart_wheelchair4}.
The system enables autonomous robot navigation within the hospital by utilizing color-coded wayfinding lanes pre-installed on the hospital floor.
To achieve this goal, the robot platform includes a mission computer and multiple camera sensors strategically placed to monitor the robot's front view and capture the wayfinding lane with a specific color on the ground. 
A dedicated motion control algorithm is also introduced to facilitate autonomous driving.
To demonstrate the capabilities of our novel system, we conducted a real-world experiment whose results are included in this study.

This approach can potentially improve patient mobility and independence within hospitals and alleviate the workload of hospital staff responsible for patient transportation.

\subsection{Related works: EPW designs}



Currently, there are numerous research outcomes and commercial products regarding EPWs.
Despite this, previous studies have mainly focused on hardware design not far from the original shape of a manual wheelchair, with limited research conducted on a towing robot capable of docking with a wheelchair \cite{docking_robot}. 
Therefore, in this subsection, we aim to scrutinize the existing wheelchair propulsion methods and identify a suitable method for our scenario. 


Existing EPW designs can be classified by their type of propulsion mechanism.
The most common design involves a pair of circular rubber main tire wheels with electric motors attached, similar to conventional manual wheelchairs.
This approach can further be categorized into two types: (i) manual wheelchairs with detachable electric motorization equipment \cite{todo_drive} and (ii) wheelchairs designed as electric from inception \cite{conventional_wheelchair}.
The former involves attaching a wheeled motor to a specific part of the main wheel rim of a manual wheelchair to create torque for wheel rotation.
The benefit of this approach is its cost-effectiveness in converting manual wheelchairs into motorized ones.
Using this approach,  an EPW can also revert back to a manual wheelchair when necessary.
In the second type, motors are inherently designed to be connected to the drive axle of the main wheels, providing power to generate the necessary torque for motion.
This type of system is designed to be further optimized for electric propulsion, resulting in benefits such as durability and longer range; however, it has a relatively higher cost.
Steering the wheelchair in both cases is primarily accomplished through the difference in rotational speeds between the main wheels.
Caster wheels that ensure unrestricted rotation of both front wheels are utilized to prevent kinematic constraints.

Innovative designs exist to overcome the limitations of conventional approaches.
One example is called ``Smart Drive''; the motorized auxiliary power module is installed underneath the wheelchair \cite{smartdrive}.
Unlike the other add-on types, the system has its own motorized wheels that come in direct contact with the ground.
The auxiliary wheels consist of a pair of omniwheels, which can generate both linear and rotational motion by utilizing the difference in angular velocity between the two omniwheels.
This system consists of a single assembly; therefore, the attach and detach processes become simpler than conventional detachable motorization toolkits.
Another innovative example is the ``PiiMo'' platform, developed by Panasonic, which is a built-in motor wheelchair but replaces the pair of front caster wheels with omniwheels \cite{piimo}.
The freely rotating front omniwheels do not provide any kinematic constraints in the translational or rotational motion of the platform, similar to caster wheels.
However, the system becomes more durable than caster-wheel front wheels since the front wheels do not change rotating direction.


Based on the above investigations, we can see that the most suitable approach among the existing methods for electrifying manual wheelchairs per the user's short-term needs is the detachable type. 
Nonetheless, this type of equipment does not provide immediate electrification because it is difficult to attach and detach.
Additionally, it is even more challenging to integrate the electrification equipment with a wheelchair autonomously.
Therefore, such cases reinforce the rationale for developing a robot platform that can automatically connect to a wheelchair, tow it, and perform autonomous detachment after arrival.




\section{Platform Design}
This section outlines the design process of a proposed robotic platform capable of autonomously driving a manual wheelchair. 
Initially, the requirements for designing a platform optimized for hospital environments are analyzed, and a hardware concept that reflects these requirements is presented. 
Subsequently, the drive system of the robot, the docking mechanism for coupling between the wheelchair and the robot, and the selection and placement of sensors for autonomous driving are described in detail.

\subsection{Platform requirements}
Selecting proper hardware mechanisms capable of achieving a minimum volume and weight is essential to strike a balance between efficient robot operation and reducing obstruction in hospital environments. 
Furthermore, robots must be capable of self-driving and smoothly towing wheelchairs to their destination. 
Therefore, the proposed robot platform needs sufficient controllability degrees of freedom (cDoFs) in both standalone and towing modes.

Here, we introduce a coupling mechanism between the manual wheelchair and the robot and select a driving mechanism that provides sufficient cDOFs. 
The coupling mechanism should be lightweight, compact, and easy to attach and detach from manual wheelchairs with diverse designs. 
The robot's driving mechanism must be reliable and ensure smooth control of the speed and direction of the robot's motion, not only when driving by itself but also when towing the wheelchair.

\subsubsection{Coupling mechanism}
In coupling the wheelchair and robot, there are two major towing strategies to consider.
The first strategy is ``Total lifting,'' in which the entire wheelchair is placed on top of the robot and only the robot's wheels are used for autonomous driving.
The second strategy is ``Dock-and-tow,'' where the robot is connected to a part of the wheelchair and tows the system.
In the first method, the robot's kinematic driving characteristics do not change because the wheelchair wheels do not come into contact with the ground.
However, since the robot has to lift the entire wheelchair, the robot requires a larger ground area to occupy any additional lifting mechanisms to lift the wheelchair, which is a disadvantage for hospital operations.
On the other hand, the second method involves connecting the robot to a part of the wheelchair while some of the wheels remain in contact with the ground, which may cause motion constraints.
However, since the robot only needs to provide the necessary horizontal towing force by coupling only with a portion of the wheelchair structure, the weight and volume of the robot itself can be minimized.

In this study, we prioritized the suitability for operation in a narrow hospital environment, and therefore adopted the Dock-and-tow method.

\subsubsection{Drive system}
Next, we discuss the selection of the drive unit.
As mentioned earlier, the proposed robot platform has two modes of operation: a standalone mode and a towing mode.

In the standalone mode, traveling along the narrowest cross-section of the platform is required to avoid obstructing the movement of medical staff and patients. 
The turning radius should also be minimized for the same reason. 
In the towing mode, the robot should have a minimum turning radius identical to the standalone mode, even with the mechanical constraints caused by the attached wheelchair, and should enable agile driving.

Ultimately, we adopted a mechanum wheel-based drive system with holonomic driving characteristics \cite{mecanum_wheelchair} to meet these goals in both operation modes.

\subsection{Overall design}
\begin{figure}[t]
    \centering
    \includegraphics[width=1\columnwidth]{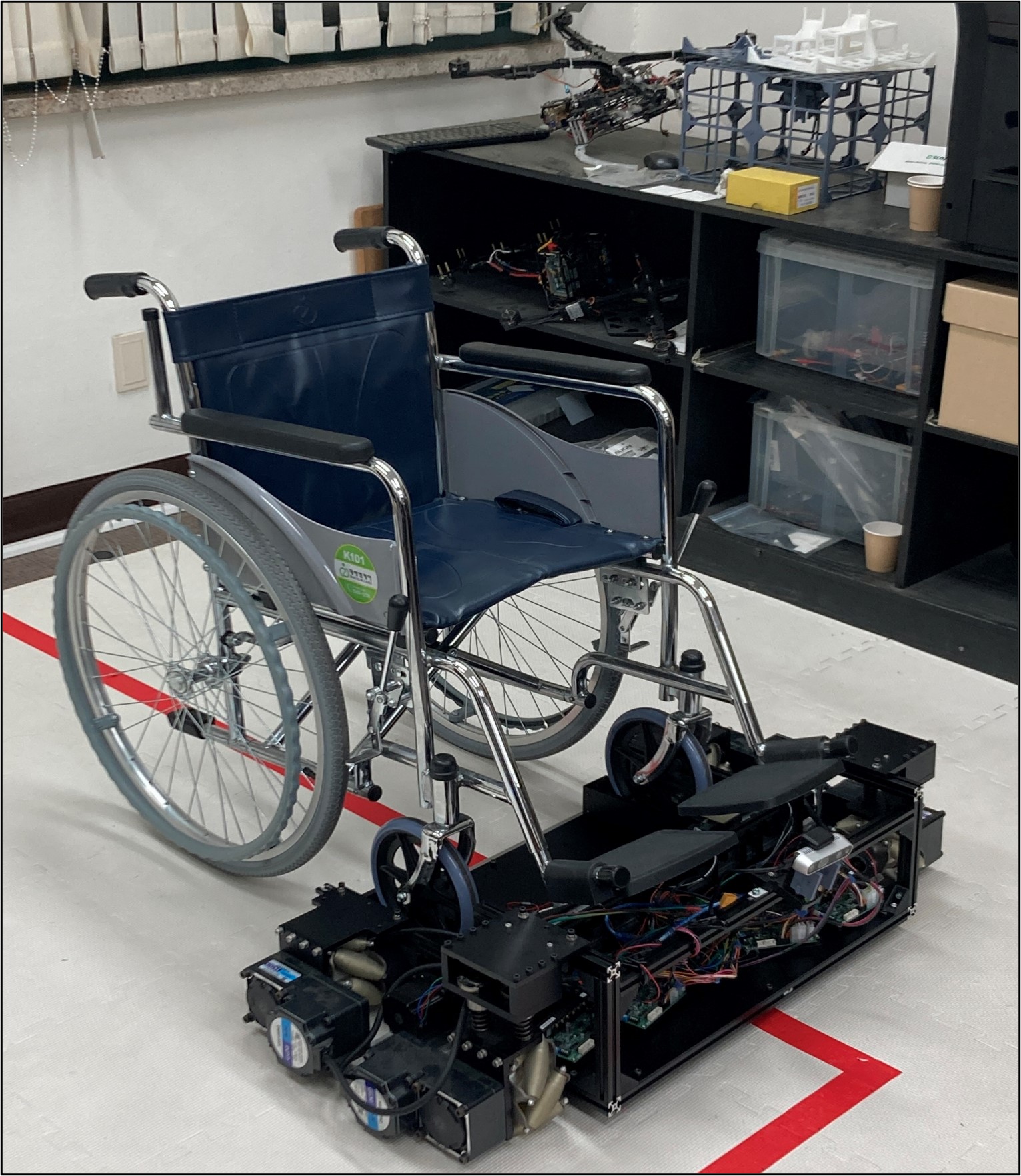}
    \caption{Configuration of the proposed wheelchair towing robot system. The proposed configuration involves coupling with only the front pair of caster wheels of a manual wheelchair while the main wheels remain in contact with the ground. This achieves a smaller size for the robot and enables lightweight construction by handling only the load imposed by the caster wheels.}
    \label{fig:wheelchair-robot_assembly}
\end{figure}
The design of the proposed wheelchair towing robot with a manual wheelchair attached is shown in Figure \ref{fig:wheelchair-robot_assembly}.
To minimize the volume and weight, the platform is designed to be coupled only with the front caster wheel of the wheelchair.
Once the coupling between the robot and the manual wheelchair is complete, the relative translational and rotational motion between the wheelchair and the robot becomes constrained. Therefore we can consider the entire assembly as a single rigid body while the coupling mechanism is activated.

In our design, the main wheel of the wheelchair remains in contact with the ground even after the docking is complete. 
Because of this, the main wheel generates kinematic motion constraints during motion, resulting in the assembled system exhibiting similar driving characteristics to a typical front-wheel drive automobile.
Conversely, during standalone operation, the towing robot has holonomic driving characteristics that are identical to a general mechanum wheel robot, where it can control two-dimensional translational motions and single-dimensional rotational motions independently.

\subsubsection{Mecanum wheel drive unit}

\begin{figure}[t]
    \centering
    \includegraphics[width=1\columnwidth]{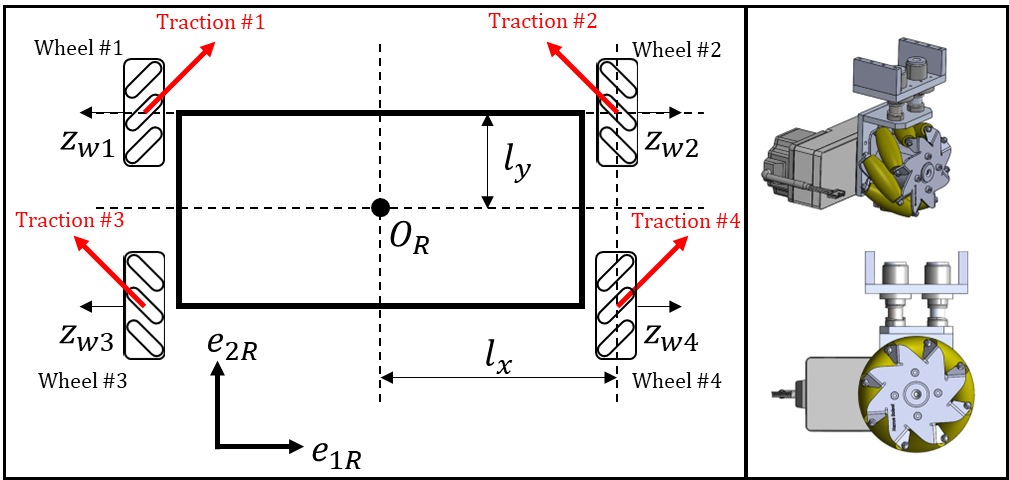}
    \caption{Schematic of a robot system with four mecanum wheels. (Left) Each mecanum wheel has auxiliary wheels installed diagonally on its periphery, generating a diagonal driving force for each wheel. By combining the rotation of the four wheels, the two-dimensional translational motion and yaw rotation of the platform can be independently controlled without slipping on any wheel. (Right) The shape of each mecanum wheel. The auxiliary wheels are oval, resulting in a perfect circular periphery when observing the assembly from the direction of the axle. This enables the mecanum wheels to achieve smooth, vibration-free motion.}
    \label{fig:schematics-solo}
\end{figure}

Figure \ref{fig:schematics-solo} shows the towing robot configuration with the mecanum wheel drive units installed on the edge of each corner of the platform.
The mecanum wheel-based drive system typically consists of four mecanum wheels, as does our platform.
The mecanum wheel drive primarily comprises an in-wheel motor system in which the wheel and motor are integrated into a single unit. 
Still, it does not allow steering of the rotational axis of the wheel. 
Therefore, the controller drives the platform motion by combining the individual rotation speed of the motor to make the overall system behave as desired.

Each mecanum wheel has multiple freely rotating auxiliary wheels along the rim.
As shown in Figure \ref{fig:schematics-solo}, the auxiliary wheel has an oval shape (yellow parts in the figure), allowing the outermost of the entire wheel to maintain a circular shape, enabling smooth driving on a flat surface.
The mecanum wheel design allows only the auxiliary wheels to contact the ground.
Since the auxiliary wheels are attached diagonally, only the component of the traction force in the direction of the auxiliary wheel axis remains.
Then, each wheel generates a driving force in the direction of the auxiliary wheel axis.
With four wheels attached to a robot, two degrees of freedom in translational and one degree of freedom in rotational motion can be achieved by combining the traction force vectors of each mecanum wheel.

Under the assumption of no-slip conditions, it is well known that the relationship between the body motion and the wheel speeds of each wheel is as follows:
\begin{equation}
    \mathbf{\Omega}_w=M\ {^{R}\dot{\mathbf{x}}}_r,\ M=\frac{1}{R_{wheel}}
    \begin{bmatrix}
    1   &   1   &   -(l_x+l_y)\\
    -1  &   1   &   (l_x+l_y)\\
    -1  &   1   &   -(l_x+l_y)\\
    1   &   1   &   (l_x+l_y)
    \end{bmatrix}.
    \label{eq:mecanum_allocation}
\end{equation}
The symbol $\mathbf{\Omega}_w=[\omega_1\ \omega_2\ \omega_3\ \omega_4]^T\in\mathbb{R}^4$ represents wheel rotation speed along the $z_{wi}$ axis, as shown in Figure \ref{fig:schematics-solo} for each $\omega_i$, $R_{wheel}$ is a mecanum wheel diameter, $l_{\{x,y\}}$ represents the distance of the wheel from the platform center, and $^{R}\dot{\mathbf{x}}_r=[^{R}\dot{x}_r\ ^{R}\dot{y}_r\ ^{R}\omega_r]^T\in\mathbb{R}^3$ represents translational velocity and rotational speed concerning the robot's body frame $\mathcal{F}_R=\{O_R;e_{1R}\ e_{2R}\}$ \cite{mecanum}.
This relationship allows us to generate the desired rotation speed of each wheel for controlling the robot's overall motion.

\subsubsection{Docking mechanism}
\begin{figure}[t]
    \centering
    \includegraphics[width=1\columnwidth]{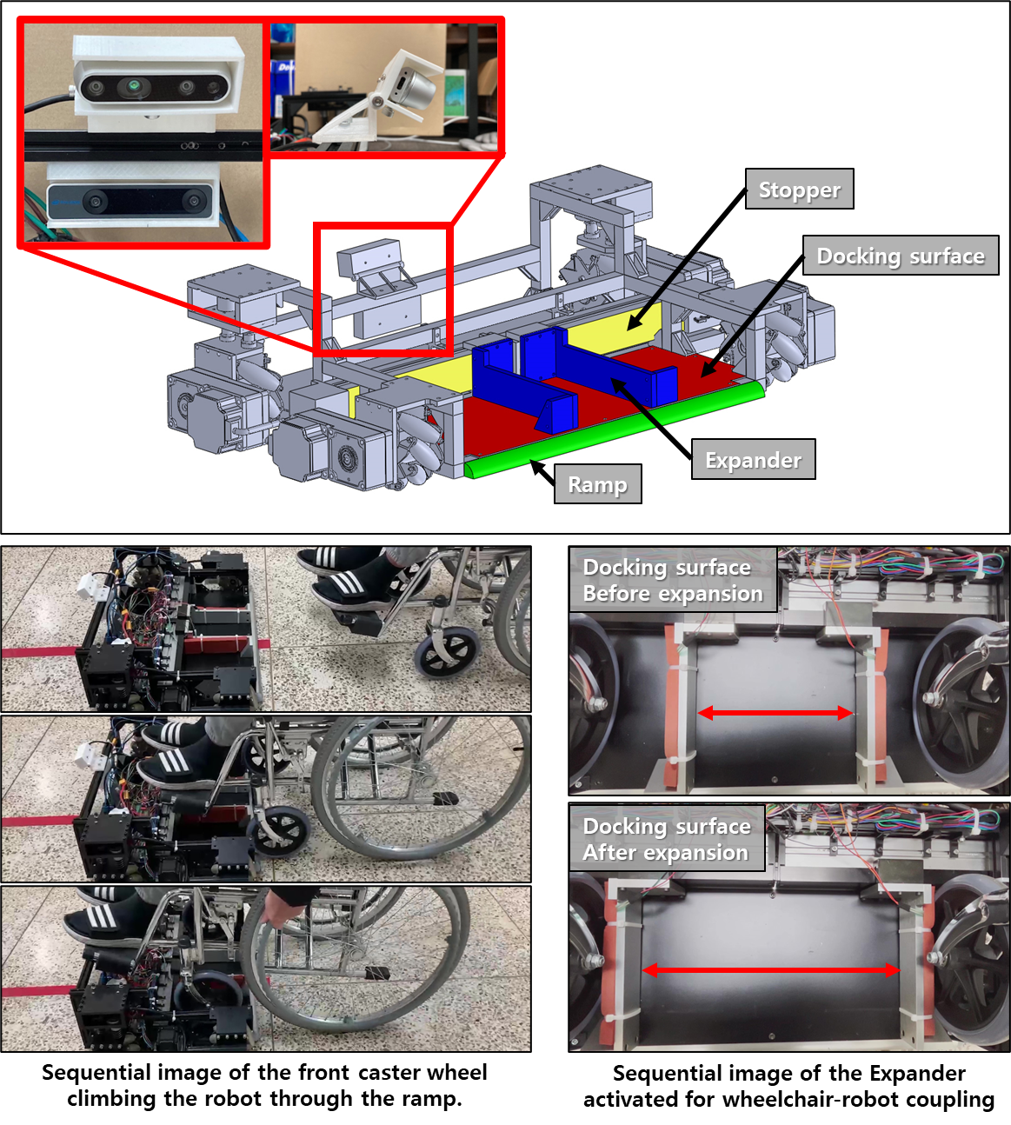}
    \caption{(Top) Design for a wheelchair towing robot. The robot is equipped with a mechanism for docking with a wheelchair and has two cameras - Intel T265 and D435i - located at the front. The D435i camera is tilted downwards to capture the wayfinding lanes on the floor. (Bottom Left) The sequence of a manual wheelchair being boarded onto a robot. When a manual wheelchair enters the docking surface, its caster wheels are lifted onto the surface via a ramp and continue to move forward until stopped by a stopper. (Bottom Right) The process of coupling the caster wheels with the robot. The Expander expands sideways to align the caster wheels and pushes them left and right. The rubber pads attached to the Expander deform as they push the caster wheels, making contact with the rim and spoke of the caster wheel and effectively transmitting the force between the robot and the wheelchair. }
    \label{fig:docking_mechanism}
\end{figure}
A new docking mechanism is devised to accommodate the varying dimensions of manual wheelchairs across different manufacturers. 
Figure \ref{fig:docking_mechanism} shows the proposed wheelchair docking system.
The new docking system consists of a friction-based coupling system that connects the robot to the front caster wheels of the wheelchair.
Initially, the wheelchair with the patient aboard approaches in front of the towing robot.
The front caster wheels of the wheelchair then move up the robot platform along a ramp, and the wheelchair is brought to a stop by a stopper in front of the caster wheels. 
Once the caster wheels are located at the designated position, a pair of bar-shaped structures with deformable rubber pads attached lies between the wheels, named the ``Expander,'' which expands laterally opposite to each other through a linear actuator.
The height of the Expander is restricted to be lower than the axis of the caster wheels, allowing the rubber pads to make direct contact with the spoke and rim of the front caster wheel structures, deforming and conforming once they make contact.
This deformation allows the towing force to be well transmitted in the forward direction of the wheelchair.
Load cells are inside the rubber pads, and when the average load value of the two bars exceeds a certain level, the linear guide rail stops and is fixed.
The Expander ultimately imposes two degrees of freedom constraints on translational motion and one degree of freedom in rotation between the robot and the wheelchair, thereby restraining any relative motion between them.

\subsection{Sensor installation}
Figure \ref{fig:docking_mechanism} also shows the sensors installed on the robot platform.
The system has two sensors for autonomous driving: Intel's T265 motion tracking camera \cite{t265} and D435i depth camera \cite{d435}.
The T265 motion tracking camera provides the current pose of the robot based on the global frame $\mathcal{F}_G=\{O_G; e_{1G}\ e_{2G}\}$.
The D435i camera provides depth information for each pixel along with an RGB image and is installed considering the future obstacle avoidance path planning mission.
In this study, only the RGB image from the D435i camera is utilized for achieving wayfinding lane detection.
The D435i camera is installed at a downward angle to effectively capture the ground image.

\section{Autonomous driving algorithm}
Our goal in performing an autonomous driving task is to extract specific color-coded wayfinding lanes pre-installed on hospital floors and enable the autonomous driving algorithm to track them.
Therefore, in this section, we introduce the following details:
\begin{itemize}
    \item A method for extracting specific colored wayfinding lanes using a camera mounted on the robot and generating a motion plan.
    \item A control algorithm for moving the robot along the extracted path, considering motion constraints.
\end{itemize}

\subsection{Motion planning}
\begin{figure}[t]
    \centering
    \includegraphics[width=1\columnwidth]{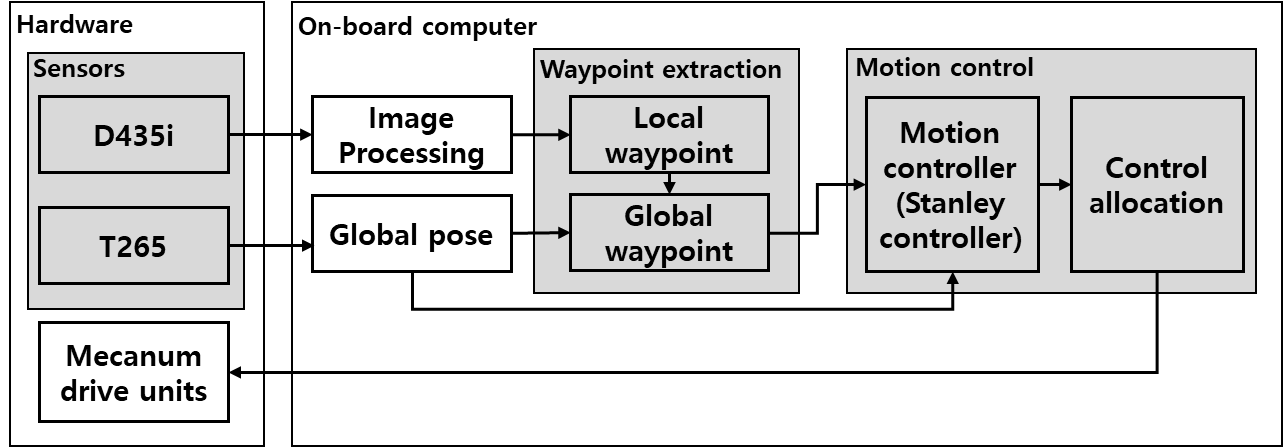}
    \caption{Overview of the wayfinding lane tracing system.}
    \label{fig:motion_planning_overview}
\end{figure}
\begin{figure*}[t]
    \centering
    \includegraphics[width=1\textwidth]{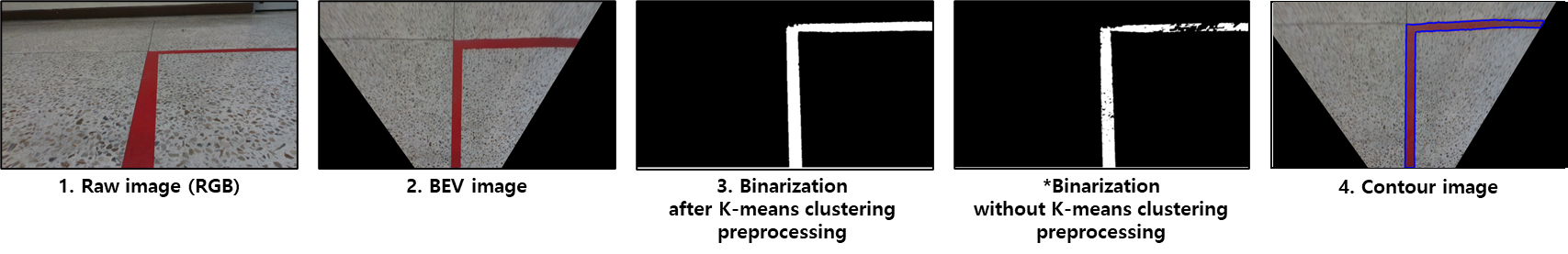}
    \caption{Processing sequence of RGB camera images. When the camera captures a new RGB image (Step 1), the image is transformed into a bird's-eye view (BEV) (Step 2). Then, through K-means clustering and bilinearization processes, only the desired lanes are detected (Step 3). The detected lanes may not be smooth if K-means clustering is not performed due to effects such as lighting glare. Finally, contours are detected (Step 4).}
    \label{fig:image_processing}
\end{figure*}

Motion planning for ground-based robots involves using optimization-based and search-based methods \cite{planning}.
The choice between these methods depends on the availability of information about the entire operation space.
In our study, we only had access to information about the robot's current pose estimate and local camera image;
thus, we employed a search-based method for our local motion planning.

To achieve a wayfinding lane tracing mission, we can consider utilizing conventional techniques such as visual servoing to control the detected lane to be located in the center of the camera image, similar to that of the lane-tracing robot.
However, in situations with vertical intersections or winding lanes, the lane image can deviate from the camera angle due to sudden changes in vehicle motion, making it difficult to perform stable lane tracing.


To tackle this problem, we employed a waypoint tracking method which generates a series of equidistant waypoints along the detected lane.
These acquired waypoints are consecutively assigned as the target position until the next image update. 
This technique provides a more reliable driving experience by ensuring path following until all the waypoints derived from the last detected image are utilized, even if multiple image updates in a particular driving scenario fail to detect a lane.

Figure \ref{fig:motion_planning_overview} shows an overview of the wayfinding lane tracing system.
First, the system acquires the forward RGB ground image through the D435i camera.
Then, the image is processed, and the waypoints with respect to the global frame are extracted.
The motion controller receives a set of waypoints and controls the robot's movement to pass through the waypoints sequentially, starting from the closest waypoint. 
This process continues until a new set of waypoints is extracted from the next camera image.

Below, we introduce the motion planning processes, including image processing and waypoint extraction and updating.

\subsubsection{Image processing}
Figure \ref{fig:image_processing} illustrates the image processing sequence. 
Our goal in this process is to extract global trajectory waypoints from the camera image. 

\begin{figure}[t]
    \centering
    \includegraphics[width=1\columnwidth]{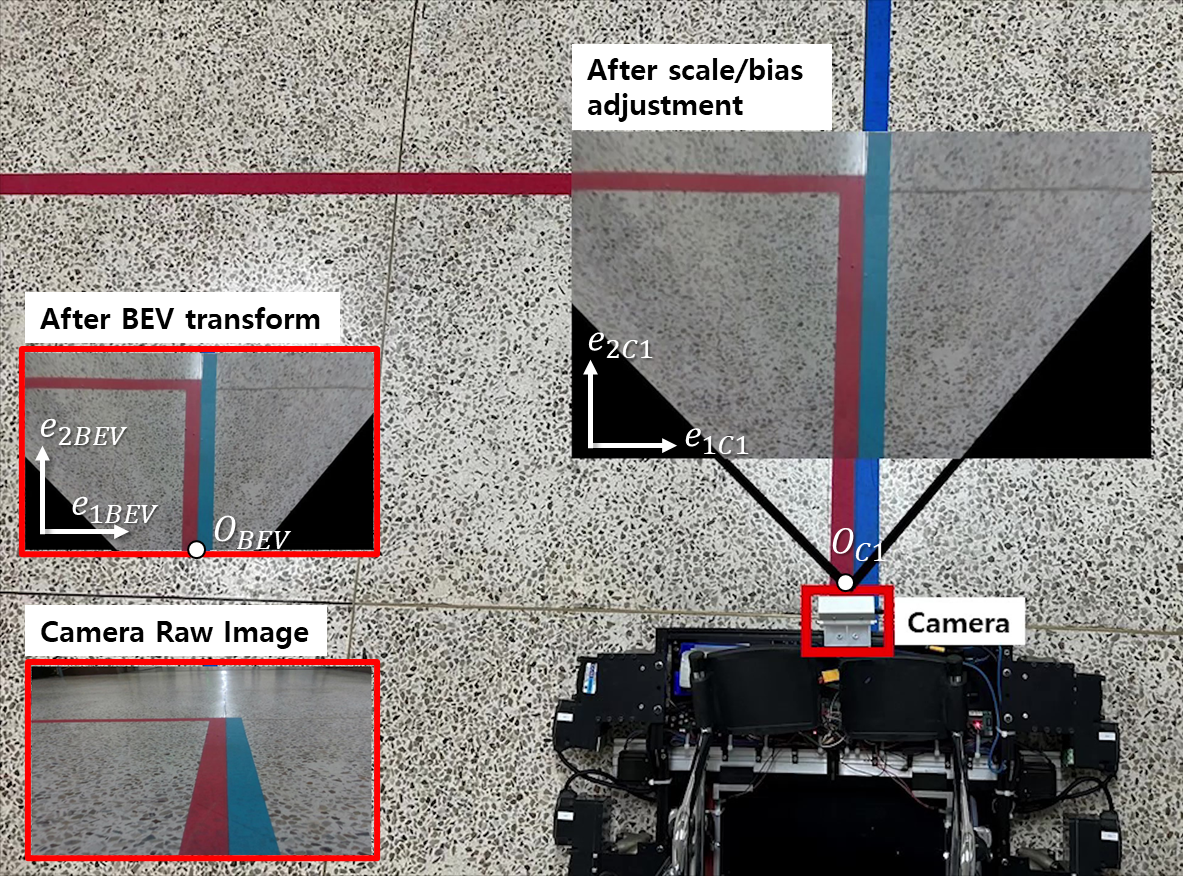}
    \caption{Process of BEV image transformation. First, the raw RGB image is transformed into a bird's-eye view (BEV) image. Then, the scale and bias are adjusted. After completing the transformation, the positions of the pixels in the image are aligned with the actual positions of objects relative to the $\mathcal{F}_{C1}$ frame.}
    \label{fig:perspective_transform_sequence}
\end{figure}

Once the floor image is received, the first step is to obtain the real-world position information of each pixel in a captured image.
With this process, we can extract the global position of waypoints along the wayfinding lane in the camera image, which will be utilized as target position values for the motion controller.
The easiest way to achieve this goal is to convert the perspective of the camera image into a bird's-eye view (BEV).

Figure \ref{fig:perspective_transform_sequence} depicts the process of BEV transformation. 
Once the raw RGB image is captured from the camera, it is transformed into the BEV.
After the BEV transformation, the pixel coordinates of the image follow the coordinates of the new image frame, $\mathcal{F}_{BEV}=\{O_{BEV};e_{1BEV}\ e_{2BEV}\}$. 
The image can then be converted to a real-world frame $\mathcal{F}_{C1}=\{O_{C1};e_{1C1}\ e_{2C2}\}$ by adjusting the scale and bias of the image. 
Once the conversion is complete, the position of each pixel in the image becomes the same as the actual position of the feature with respect to the real-world camera frame.

Next, we extract a specific lane with the designated color from the BEV images.
Our strategy is to detect lanes based on color information.
However, uncertainties arise due to various factors in the surrounding environment, such as glare from lighting, shadows from nearby foreign objects, or uneven lighting conditions, which cause color distortion or failure to detect lanes in certain areas.
To address this problem, we utilized the K-means-based color quantization technique \cite{image_quantization}. 
This technique leverages the ability of the K-means clustering method to group similar colored pixels without requiring separate training processes \cite{kmeans}, allowing to unify of the colors within each group. 
As a result, the color contrast of the image became more pronounced, emphasizing the distinction between color groups.
We tuned the number of color regions $k$ heuristically to suit the driving environment, specifically based on the colors used for lane makings.
We then set it as the initial input value to cluster color regions more robustly than before.

Through K-means clustering, the colors in the camera image are clustered, and the contrast between colors becomes more distinct.
However, certain color lanes in the image still cannot be detected as a single color criterion; therefore, specifying the range of colors and recognizing all colors within that range as lanes are necessary.
Since it is difficult to specify the range of colors in RGB-based color representation, which combines three colors to represent a final color, we transformed the color representation to HSV representation, which represents colors in hue, saturation, and value. 
Upon performing HSV transformation, a calibration process is performed to define the range of the hue parameter, which signifies color. 
This enables us to generate a binary image by assigning a ``1'' value to the designated pixels and a ``0'' to those remaining.

Finally, extraction of the contours of the wayfinding lane from the binarized image is performed.
Subsequently, we evenly distribute waypoints within the contours by following the later methodology.



\subsubsection{Waypoint extraction/update}
Our next task is to extract a series of waypoints along the contour from the final result of the previous image processing sequence.
The image segments of wayfinding lanes in hospitals can generally be categorized into two types: regular lane segments and intersection lane segments.
For regular lanes, a straight or curved lane segment is detected on the BEV-transformed camera image, and a series of waypoint arrays along the lane can be extracted.
However, this process may not work well with intersection lane segments; this, additional considerations are required.

\begin{figure}[t]
    \centering
    \includegraphics[width=1\columnwidth]{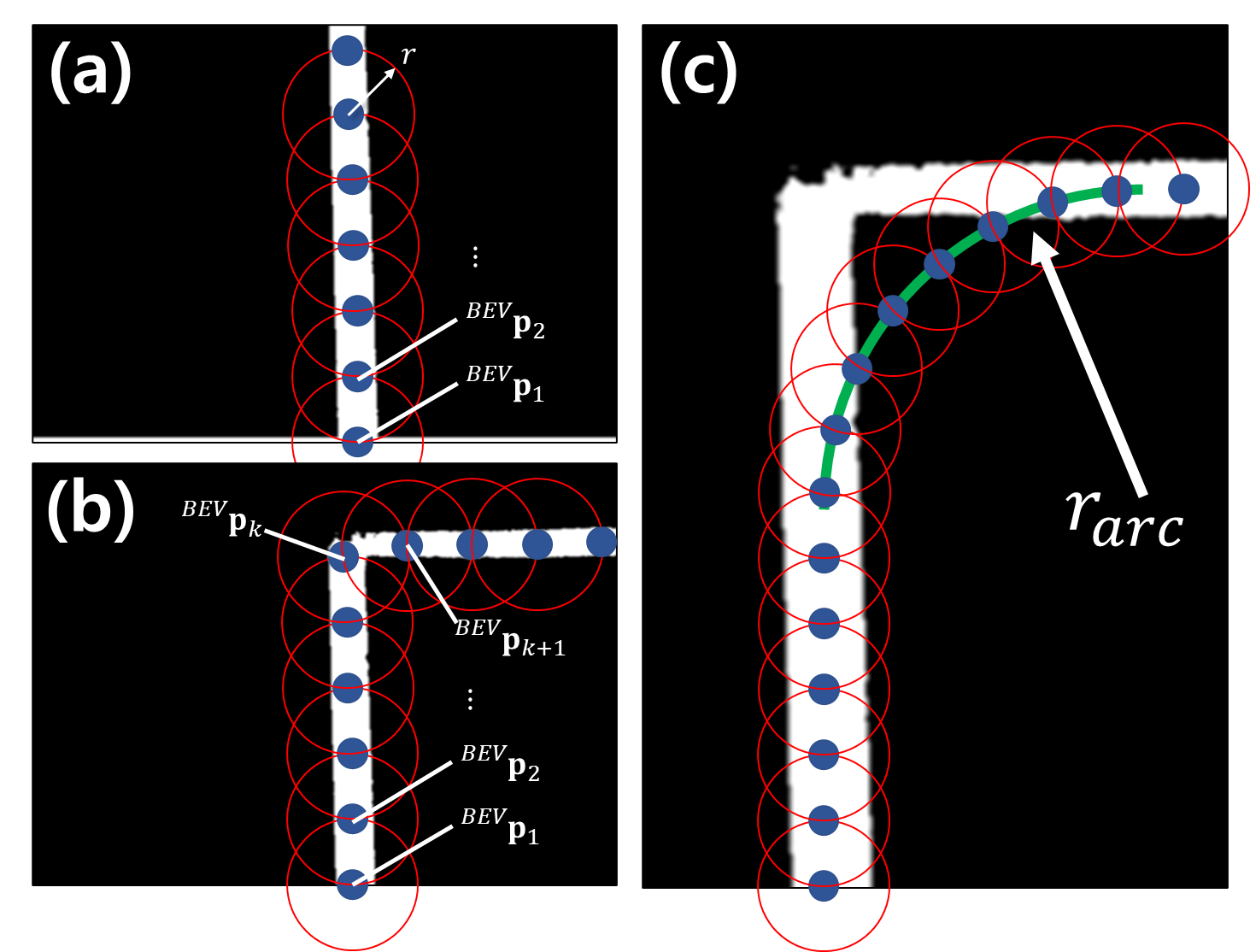}
    \caption{The process of obtaining a waypoint series from a single RGB camera image. (a) General case. The lowest central point of the wayfinding lane is designated as the starting point. A circle with radius $r$ is drawn, and the intersection point between the circle and the wayfinding lane is designated as a new waypoint. (b) Vertical intersection. A virtual arc is created that is tangent to the detected wayfinding lane, and waypoints are detected along the virtual arc. (c) Illustration describing the overall results of the waypoint detection.}
    \label{fig:waypoint_extraction}
\end{figure}

Figure \ref{fig:waypoint_extraction}-(a) illustrates the process of extracting waypoints in a general situation.
In general cases, the first waypoint $^{BEV}\mathbf{p}_1=[x_1\ y_1]^T\in\mathbb{R}^{2}$ inside the BEV camera image is determined by identifying the pixel in the lowermost row of the image that belongs to the contour and designating it as the starting point.
Then, a circle centered on this pixel with radius $r$ is drawn, and the upper intersection between the circle and the center of contours is selected as the second waypoint $^{BEV}\mathbf{p}_2$.
Repeating this process can extract a waypoint array $^{BEV}\mathbf{P}=[^{BEV}\mathbf{p}_1\ ^{BEV}\mathbf{p}_2\ \cdots\ ^{BEV}\mathbf{p}_i\ \cdots\ ^{BEV}\mathbf{p}_N]^T\in\mathbb{R}^{N\times2}$ with $N$ number of waypoints along the contour at a uniform distance $r$, regardless of whether the contour is a straight line or a curve.
The number $N$ is determined based on the captured contour and the radius $r$.

For vertical intersections, if the waypoints are extracted as described above, the driving system will demand sudden and extremely rapid vehicle rotations. This is because the placement direction of the waypoints changes rapidly at the intersection, similar to the $^{BEV}\mathbf{p}_k$ and $^{BEV}\mathbf{p}_{k+1}$ waypoints shown in Figure \ref{fig:waypoint_extraction}-(b).
To overcome this issue, a virtual smooth waypoint series is created, as shown in Figure \ref{fig:waypoint_extraction}-(c), to ensure stable driving.
Our strategy is to generate a virtual arc (green solid arc in the figure), a part of the circle with radius $r_{arc}$, tangential to the two line segments that form the intersection: then, to extract waypoints along that arc. 
Once the intersection is detected and the arc is generated, the camera stops the image update, and the vehicle follows the set of waypoints extracted from the very last image.
The waypoint extraction process resumes when the camera recognizes a straight line without an intersection.
This is because generation of the virtual arc trajectory occurs prior to the disappearance of the intersection point from the camera image.
The continuous execution of the conventional waypoint extraction process can yield discordant waypoints that compel the robot to relocate immediately to the lane center rather than following the virtual arc lane.

Once the waypoints are extracted from the image, the final task is to calculate the position of each waypoint with respect to the global frame $\mathcal{F}_G$.
First, we transform the image plane-based position data into data with respect to the D435i camera frame $\mathcal{F}_{C1}$ as follows:
\begin{equation}
    ^{C1}\mathbf{p}_i=s\ ^{BEV}\mathbf{p}_i+\mathbf{b}.
\end{equation}
Here, $s\in\mathbb{R}$ is for adjusting the scale and $\mathbf{b}\in\mathbb{R}^{2\times1}$ is for adjusting the bias, where both parameters are constant.
Afterwards, we define a homogeneous transformation-based representation $^{C1}\tilde{\mathbf{p}}_i\in SE(2)$, where
\begin{equation*}
    ^{C1}\tilde{\mathbf{p}}_i=
    \begin{bmatrix}
        I_{2\times2} & ^{C1}\mathbf{p}_i\\
        0_{1\times2} & 1
    \end{bmatrix}.
\end{equation*}
If we define $^{C2}T_{C1}\in SE(2)$ as the pose of the D435i with respect to the T265 camera frame ($C2$ frame), and $^{G}T_{C2}\in SE(2)$ as the current pose of the T265 camera, we can calculate the position information of the extracted waypoint in the $\mathcal{F}_G$ frame on the image plane through the following equation:
\begin{equation}
    ^G\tilde{\mathbf{p}}_i={^GT_{C2}}{^{C2}T_{C1}}{^{C1}\tilde{\mathbf{p}}_i}.
\end{equation}
Here, $^{C2}T_{C1}$ is a fixed value, and $^{G}T_{C2}$ is derived from the pose estimation of T265 camera as follows
\begin{equation}
    ^{G}T_{C2}=
    \begin{bmatrix}
        \cos(^G\psi) & -\sin(^G\psi) & ^Gx_r\\
        \sin(^G\psi) & \cos(^G\psi) & ^Gy_r\\
        0 & 0 & 1
    \end{bmatrix},
    \label{eq:gTc2}
\end{equation}
where $^G\psi$, $^Gx_r$, and $^Gy_r$ are the robot's orientation and position with respect to the global origin.

\subsection{Motion control}
The control methodology for the standalone drive mode is consistent with that of a typical mecanum wheel robot \cite{mecanum2}. 
Therefore, in this section we focus solely on the control strategy for the towing mode.
We first introduce a platform motion control strategy considering motion constraints when combining a wheelchair.
Next, we introduce a Stanley motion control technique for tracing an array of waypoints \cite{stanley}.
Finally, we describe the rules for updating waypoints within a waypoint array $^{BEV}\mathbf{P}$.

\subsubsection{Platform motion control considering motion constraint}
Once the coupling completes, the relative position and orientation between the robot and the wheelchair become fixed, and the entire system can be considered a single rigid body.
In this state, the main wheels of the wheelchair come into contact with the ground, generating motion constraint.
Failure to consider this constraint during control may result in slipping by both the mecanum wheels and wheelchair main wheels or undesired disengagement of the coupling between the robot and the wheelchair due to excessive internal forces acting on the coupling mechanism.

\begin{figure}[t]
    \centering
    \includegraphics[width=1\columnwidth]{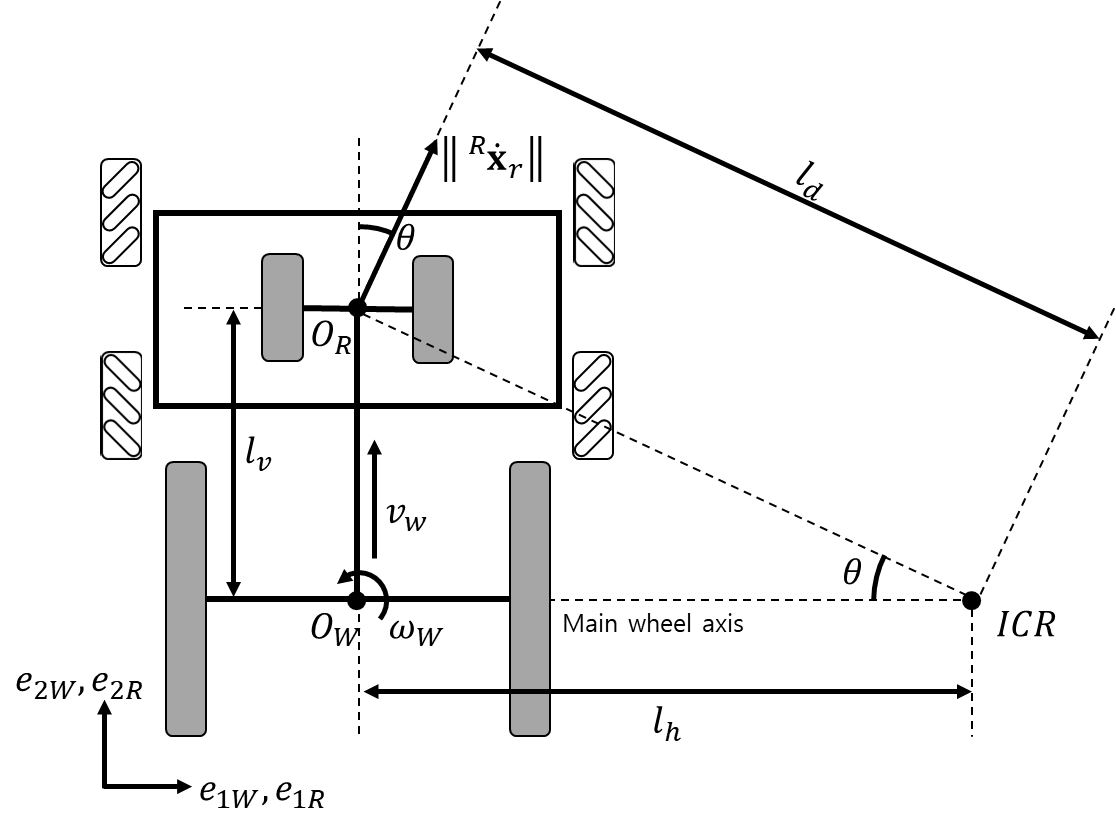}
    \caption{Schematics of the wheelchair-towing system. After wheelchair coupling, the entire system can be considered a single rigid body. The instantaneous center of rotation of the system should always be located on the axis of the main wheel of the wheelchair for non-slip motion.}
    \label{fig:coupled_schematics}
\end{figure}
Figure \ref{fig:coupled_schematics} shows the configuration of our robot system combined with a wheelchair.
To consider the kinematic constraints caused by the main wheels, the Ackerman steering geometry \cite{ackerman} is introduced.
Similar to the rear wheels of a typical front-wheel steering automobile, the main wheelchair wheels cannot be steered. 
Therefore, non-slip motion is possible only when the main wheels move vertically to the main wheel axis.
According to the Ackerman steering geometry, this vertical movement is possible only when a vehicle's instantaneous center of rotation (ICR), which is the center of rotation shared by all rigid body components, is located along the rear wheel axis.
Suppose the ICR is located along the axis. 
In that case, the rear wheels then always generate speed in a direction perpendicular to the main wheel axis, shown in the figure. 
Then there is no axis-directional translational motion of the main wheels of a wheelchair, meaning no slip occurs.

Regarding the towing robot's satisfaction with the motion constraint, the ICR of the towing robot must also be the same as that of the wheelchair.
Let $\theta$ be the angle between the main wheel axis and a line segment connecting the $O_R$ and ICR in Figure \ref{fig:coupled_schematics}.
To make the ICR of the robot identical to the wheelchair, the direction of motion must always be perpendicular to the line segment.
Also, the robot's rotation speed must be the same as the wheelchair.
When considering all the aforementioned constraints, there are two remaining controllable parameters for maneuvering the motion of the platform: the location of the ICR on the main wheel axis, which is the turning radius of the platform, and the absolute value of the robot vehicle velocity.

If we define the turning radius of the system as $l_h$, we can determine the position of the ICR by controlling this value.
Then, the remaining parameters for steering the system are $\|^R\dot{\mathbf{x}}_r\|=\sqrt{{^R\dot{x}_r^2}+{^R\dot{y}_r^2}}\in\mathbb{R}$, which is the velocity magnitude (=speed) of the robot vehicle.
To realize the motion, we need to convert $\|^R\dot{\mathbf{x}}_r\|$ and $l_h$ signals to $^R\dot{\mathbf{x}}_r$, the speed vector concerning the $\mathcal{F}_R$ frame (refer to Equation \ref{eq:mecanum_allocation}). This relationship is defined as follows:
\begin{equation}
    ^R\dot{\mathbf{x}}_{r}=
    \dfrac{\|^R\dot{\mathbf{x}}_r\|}{l_d}
    \begin{bmatrix}
        l_v sgn(l_h)\\
        l_h\\
        -sgn(l_h)
    \end{bmatrix},
    \label{eq:control_relationship}
\end{equation}
where $sgn(*)$ represents the signum function.

Equation (\ref{eq:control_relationship}) is derived with respect to the perspective of the robot frame.
By controlling the motion based on the wheelchair frame $\mathcal{F}_W$, Equation (\ref{eq:control_relationship}) can further be simplified.
Let $v_w$ and $\omega_w$ be the wheelchair's translational and rotational speeds, respectively, as represented in Figure \ref{fig:coupled_schematics}.
Then the following relationship holds:
\begin{equation}
\left\{
\begin{array}{lr}
    v_w=\|^R\dot{\mathbf{x}}_r\|\cos{\theta}\\
    \omega_w=-sgn(l_h)\dfrac{\|^R\dot{\mathbf{x}}_r\|\sin{\theta}}{l_v}
\end{array}
\right..
\label{eq:robot_to_wheelchair}
\end{equation}
Subsequently, we can obtain the following equation by applying Equation (\ref{eq:robot_to_wheelchair}) to (\ref{eq:control_relationship}) as,
\begin{equation}
^R\dot{\mathbf{x}}_{r}=
Ac_w, A=
\begin{bmatrix}
    0 & l_v\\
    1 & 0\\
    0 & 1
\end{bmatrix},
\label{eq:control_allocation}
\end{equation}
where $c_w=[\omega_w\ v_w]^T\in\mathbb{R}^{2}$.
Ultimately, we can control the translational and rotational speed of the wheelchair by using Equation (\ref{eq:control_allocation}).

\subsubsection{Stanley motion controller}
\begin{figure}[t]
    \centering
    \includegraphics[width=.9\columnwidth]{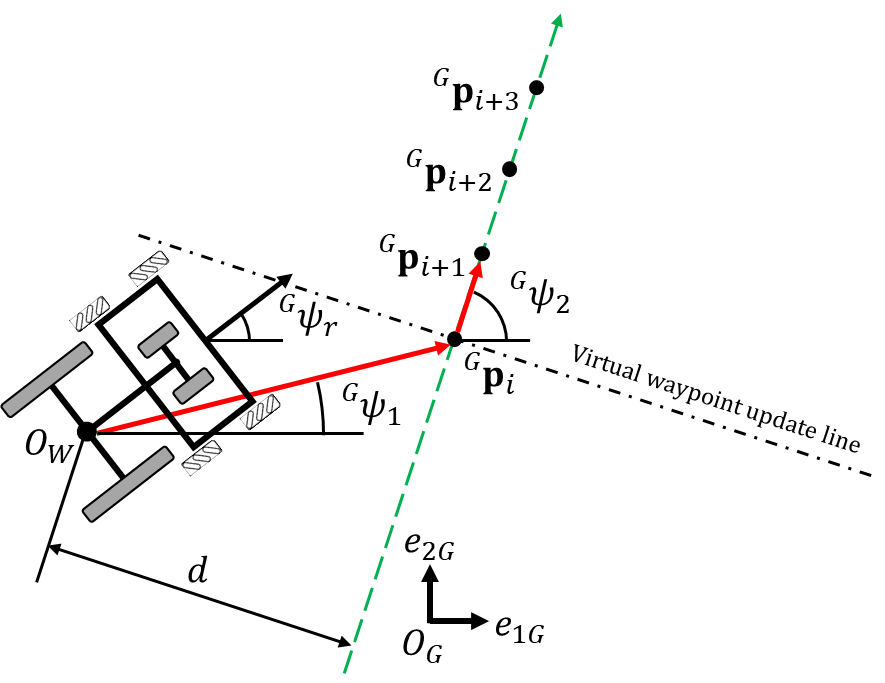}
    \caption{Operation of the Stanley motion controller. If the cross-track error $d$ is large, the robot travels toward the target waypoint ($^G\psi_1$). As $d$ decreases, the robot switches to the direction of the wayfinding lane ($^G\psi_2$).}
    \label{fig:stanley_controller}
\end{figure}

Once the array of waypoints is determined through path planning, the robot travels along the desired path by sequentially following the target waypoints.
If the robot blindly controls its movement using the ``pure pursuit'' method \cite{pure_pursuit} only to head towards the target waypoint, it can eliminate the cross-track error from the wayfinding lane. 
Still, it may not be able to stably track the wayfinding lane due to its short lookahead distance.
To overcome this problem, the Stanley motion control method is introduced \cite{stanley}. 

Figure \ref{fig:stanley_controller} shows the operation of the Stanley motion controller.
Assuming that the speed $v_w$ is constant, the desired orientation of the vehicle is as follows:
\begin{equation}
    ^G\psi_{r,des}=k(d){^G\psi_1}+\left(1-k(d)\right){^G\psi_2},
    \label{eq:stanley}
    \end{equation}
where
\begin{equation*}
    \left\{
    \begin{array}{lr}   
    k(d)=0.5\left({1+\tanh{(d)}}\right)\\
    d=\|^G\mathbf{p}_i-^G\mathbf{x}_W\|\sin{\left({^G\psi_1}-{^G\psi_2}\right)}    
    \end{array}
    \right..
\end{equation*}
$^G\psi_{r}$ represents the current heading angle of the robot in the $\mathcal{F}_G$ frame, and $^G\psi_1$ indicates the direction from the wheelchair to the current target waypoint.
$^G\psi_2$ represents the direction of the wayfinding lane, and $k(d)$ is a blending function dependent on the cross-track distance $d$.

The platform performs pure pursuit toward the waypoint if the cross-track error is large.
However, as the cross-track error decreases, the target angle $^G\psi_{r,des}$ gradually adjusts towards the direction of the wayfinding lane. 
Ultimately, when the cross-track error is eliminated, the robot travels along the lane and follows the path.

The target waypoint must be continuously updated to continue following a wayfinding lane. 
The condition for updating the target point from the $i$-th waypoint $^G\mathbf{p}_i$ to the $i+1$-th waypoint $^G\mathbf{p}_{i+1}$ is as follows:
\begin{equation*}
\lvert{^G\psi_2}-{^G\psi_1}\rvert>\dfrac{\pi}{2}.
\end{equation*}
The condition involves creating a virtual waypoint update line that penetrates the current waypoint and is perpendicular to the tangent of the wayfinding lane at that particular point. 
Subsequently, the robot updates the waypoint when it crosses this virtual line. 
This approach ensures a reliable update of waypoints, even in the presence of a residual cross-track error.

In summary, the robot platform takes the RGB images from the D435i camera and pose estimate information from the T265 camera as inputs, and outputs the $^G\dot{\mathbf{x}}_r$ data to drive the robot's mecanum wheels for autonomous driving (refer to Figure \ref{fig:motion_planning_overview}).
The overall algorithm is structured as Algorithm \ref{al:overall}.
\RestyleAlgo{ruled}
\begin{algorithm}
\SetKwInOut{Input}{Input}
\SetKwInOut{Output}{Output}
\SetKwInOut{Data}{Data}
\SetKwInOut{Params}{Params}

\caption{Driving algorithm}\label{alg:cap}
\smaller
\Data{Image data: $I_{*}$\\
            Array with $N$ waypoints: $^G\mathbf{P}$\\
        Desired robot position: $^G\mathbf{X}_{des}$\\
}
\Input{New camera image: $I_{new}$\\
        Current robot position: $^G\mathbf{X}$
}
\Output{$^R\dot{\mathbf{x}}_r$}
\Params{
        \# of waypoints to extract: $N$\\
        Hue range: $H_{min,max}$\\
        \# of color regions: $k$\\
        \Comment{for K-means clustering}
}
$i \gets 1$\;
$N \gets 10$ \Comment{Number of waypoints extracted from a single camera image}\; 
\While{$true$}{
    \While{$i\leq N$}{
        \If{$I_{new}$ received}{
            $I_1 \gets$ perspectiveTransform($I_{new}$)\;
            \Comment{to BEV}\;
            $I_2 \gets$ K-meansClustering($I_1$)\;
            $I_3 \gets$ RGB\_to\_HSV($I_2$)\;
            $I_4 \gets$ bilinearization($I_3$,$H_{min}$,$H_{max}$)\;
            $I_5 \gets$ findContour($I_4$)\;
            $^G\mathbf{P}(N) \gets$ extractWaypoints($I_5$,$N$)\;
            $i \gets 1$
        }
        $^G\mathbf{X}_{des} \gets {^G\mathbf{p}_i}$\;
        \While{$\lvert{^G\psi_2}-{^G\psi_1}\rvert<\dfrac{\pi}{2}$}{
            StanleyControl($^G\mathbf{X}_{des}$, $^G\mathbf{X}$)\;
        }
        $i\gets i+1$\;
    }
    $v_w=0$ \Comment{stop the vehicle}\;
    \If{$I_{new}$ received}{
        $i\gets1$;
    }
}
\label{al:overall}
\end{algorithm}

\begin{figure*}[t]
    \centering
    \includegraphics[width=1\textwidth]{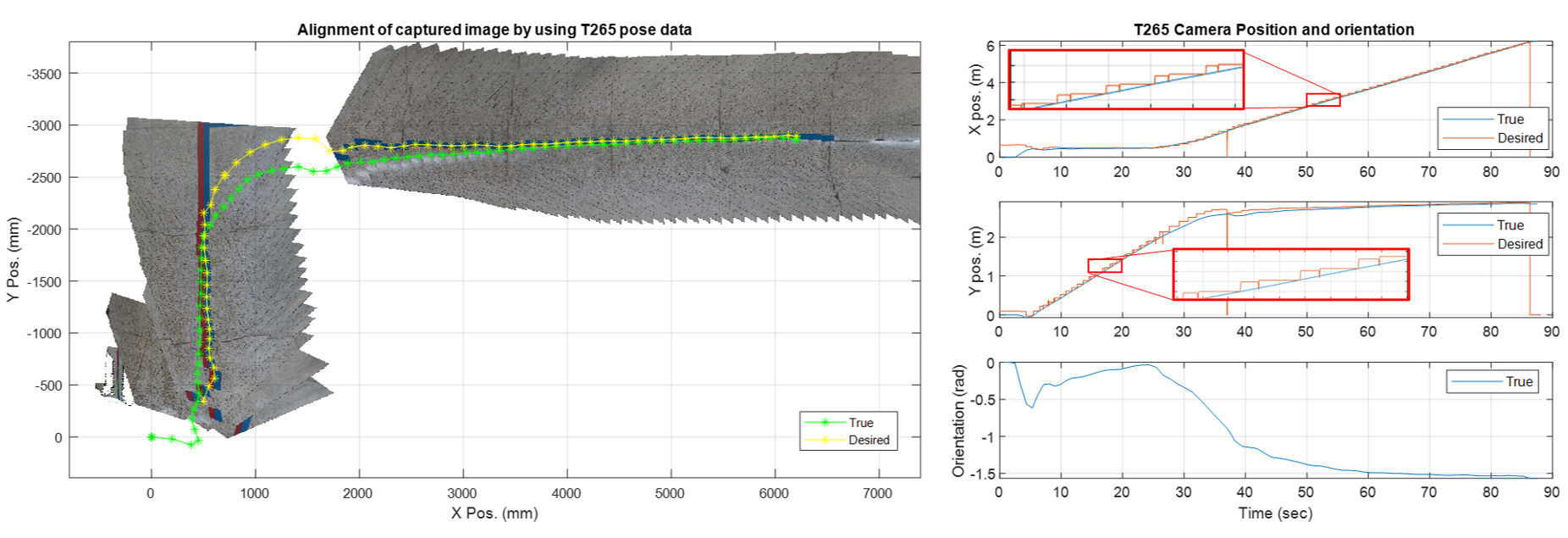}
    \caption{Wayfinding lane tracing experiment result. \textbf{[Left]} A photo with individual BEV images placed based on T265 camera position information. The desired waypoint array is aligned along the lane on the camera image, and waypoints are generated along a virtual arc trajectory at the intersection. In the beginning of the experiment, cross-track errors were introduced, but the trajectory converges to the wayfinding lane soon after the beginning by the Stanley controller. \textbf{[Right]} The robot's desired and true pose graph. A new waypoint is updated when the robot reaches the target position; thus, the desired position shows a step-like shape. The waypoint is updated before the robot reaches the desired waypoint due to new image acquisition and a new waypoint array generation. However, overall, a smooth driving trajectory can be confirmed.}
    \label{fig:experiment}
\end{figure*}

\section{Experiment}
\begin{figure}[t]
    \centering
    \includegraphics[width=1\columnwidth]{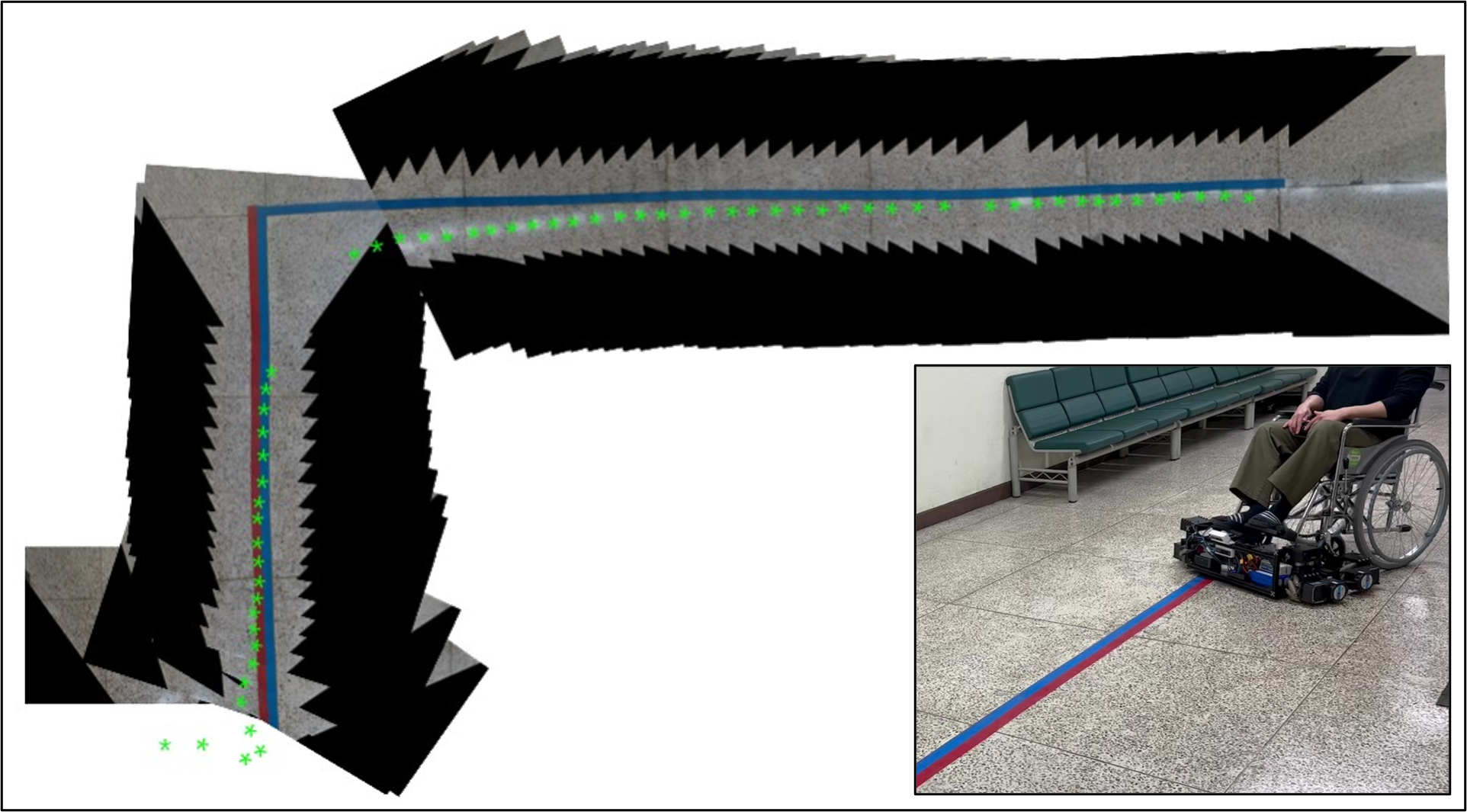}
    \caption{Bird's-eye view of the overall experiment environment and the snapshot of the experiment with patients onboard. The image is created through panoramic image composition (without sensor pose data), and the position of the robot, indicated by the green asterisk mark, is drawn based on the fact that the robot's location for each individual image frame has a fixed value.}
    \label{fig:image_stitched}
\end{figure}

An experiment validated the lane-tracking algorithm, which involves constructing the experimental environment depicted in Figure \ref{fig:image_stitched}. 
The environment consisted of two colored lines attached to the center of a corridor. 
The post-processed image in Figure \ref{fig:image_stitched} was obtained by stitching together RGB camera images captured by the robot to produce a BEV true image of the entire experimental space. 
The robot's pose information was not used during the stitching process, and morphing of individual camera images was not performed. 
Since the camera is firmly attached to the robot, the robot's position for an individual image can be determined, and the green asterisk in Figure \ref{fig:image_stitched} indicates the robot's actual position. 

The experimental results are presented in Figure \ref{fig:experiment}. 
Unlike Figure \ref{fig:image_stitched}, which depicts image stitching based on features within individual images, Figure \ref{fig:experiment} is a collection of images arranged using the robot's position and orientation information obtained from the T265 pose sensor.
The figure's desired waypoints $^G\tilde{\mathbf{p}}_i$ represent only the waypoints that are actually used for driving among the numerous waypoint arrays extracted from individual images. 
The true position represents the robot's current position derived from the T265 sensor. 
The graphs on the right-hand side of the figure show the target and actual positions of $X$ and $Y$ in the global frame, with the target position updating in a stepped pattern as the robot arrives at each waypoint. 
The third graph represents the robot's orientation, which gradually converges to a heading angle of $-\pi/2$ after rotating corresponding to the intersection from around 25 seconds to capturing a new straight line at around 37 seconds, as can be observed in the enlarged graph.

By comparing the pre- and post-intersection passage in Figure \ref{fig:experiment}, we can observe a drift in the T265-based position information due to the increased bias of the robot's pose estimation acquired by the local pose sensor. 
However, the proposed driving algorithm employs a local planning technique that extracts the target waypoint based on the on-board RGB sensor image.
This enables the robot to maintain a consistent and stable ability to track the wayfinding lane.
A full video of the experiment can be found at \url{https://www.youtube.com/watch?v=RdnuhTjrMQc}.

\section{Conclusion}
In this paper, we introduced a new wheelchair towing robot system that can selectively electrify manual wheelchairs in a typical hospital environment equipped with wayfinding lanes on the floor.
The new towing robot is designed to easily attach and detach to the front castor wheels of a manual wheelchair, providing a stable connection force that makes the entire system a single rigid body during motion.
The robot's driving mechanism utilizes mecanum wheels, allowing for free movement during independent driving and necessary towing force generation that meets all motion constraints imposed by the wheelchair during towing. This paper also presented an autonomous driving algorithm that acquires local waypoint information from forward images captured during motion to address the pose drift of the local sensor.
We introduced the design of a Stanley controller for effectively following the wayfinding lanes and presented a stable transition driving technique at the intersection course.
Finally, we validated the driving performance of the platform through a practical experiment.

At present, the robot system introduced lacks the ability to avoid obstacles. 
Given the complex nature of hospital environments, it is essential to avoid people and objects inside the hospital actively.
In future research, we plan to utilize the D435i depth camera embedded in the system to perform studies on obstacle detection and active avoidance during wayfinding lane following, as well as enabling the robot to return to its path.





\bibliography{sn-bibliography}


%

\end{document}